\documentclass[conference]{IEEEtran}
\IEEEoverridecommandlockouts
\usepackage{cite}
\usepackage[colorlinks=true, allcolors=blue]{hyperref}
\usepackage{amsmath,amssymb,amsfonts}
\usepackage{algpseudocode}
\usepackage{graphicx}
\usepackage{textcomp}
\usepackage{xcolor}
\usepackage{svg}
\usepackage{multirow}
\usepackage{array}
\usepackage{caption}
\usepackage{float}
\DeclareCaptionLabelSeparator{bluecolon}{\textcolor{blue}{: }}
\usepackage[ruled,vlined]{algorithm2e}
\def\BibTeX{{\rm B\kern-.05em{\sc i\kern-.025em b}\kern-.08em
    T\kern-.1667em\lower.7ex\hbox{E}\kern-.125emX}}
\begin{document}

\title{Time Series Anomaly Detection with CNN for Environmental Sensors in Healthcare-IoT\\
{}
\thanks{This publication has emanated from research conducted with the financial support of Science Foundation Ireland under Grant number 18/CRT/6049.}
}

\author{
\IEEEauthorblockN{Mirza Akhi Khatun}
\IEEEauthorblockA{Dept. Electronic \& Computer Engineering,\\
SFI CRT Foundations in Data Science\\
University of Limerick\\
Limerick, V94 T9PX, Ireland\\
\href{mailto:mirza.akhi@ul.ie}{\color{black}{mirza.akhi@ul.ie}}}\\
\IEEEauthorblockN{Ciarán Eising}
\IEEEauthorblockA{Dept. Electronic \& Computer Engineering\\
University of Limerick\\
Limerick, V94 T9PX, Ireland\\
\href{mailto:ciaran.eising@ul.ie}{\color{black}{ciaran.eising@ul.ie}}}
\and 
\IEEEauthorblockN{Mangolika Bhattacharya}
\IEEEauthorblockA{School of Information Technology\\
Illinois State University\\
Illinois, USA\\
\href{mailto:mbhatt1@ilstu.edu}{\color{black}{mbhatt1@ilstu.edu}}}\ \\
\IEEEauthorblockN{Lubna Luxmi Dhirani}
\IEEEauthorblockA{Dept. Electronic \& Computer Engineering\\
University of Limerick\\
Limerick, V94 T9PX, Ireland\\
\href{mailto:lubna.luxmi@ul.ie}{\color{black}{lubna.luxmi@ul.ie}}}
}
\maketitle 

\thispagestyle{plain}
\pagestyle{plain}\begin{abstract}
This research develops a new method to detect anomalies in time series data using Convolutional Neural Networks (CNNs) in healthcare-IoT. The proposed method creates a Distributed Denial of Service (DDoS) attack using an IoT network simulator, Cooja, which emulates environmental sensors such as temperature and humidity. CNNs detect anomalies in time series data, resulting in a 92\% accuracy in identifying possible attacks.


\end{abstract}

\begin{IEEEkeywords}
Healthcare-IoT (H-IoT), Anomaly Detection, DDoS Attack, Cyberattack, Wireless Sensor Network (WSN), Cooja Simulator, Contiki OS.
\end{IEEEkeywords}

\section{Introduction}
The Internet of Things (IoT) is a network of physical devices that communicate with each other through sensors, software, and connectivity \cite{10371310},\cite{dhirani2021industrial},\cite{zardari2023iot}. 


Considering that Wireless Sensor Networks (WSNs) are used to monitor the environment in health facilities, the necessity for precise and reliable data is significant.  Aligning an organization's cybersecurity measures is crucial for protecting patient privacy and safety and ensuring the continuous and efficient provision of high-quality care \cite{meagher2023cyber}. The massive influx of data makes it challenging to discern the presence of a malfunctioning sensor, environmental changes, or abrupt temperature fluctuations.
For example, patients' rooms with high humidity can create a breeding ground for bacteria, posing a heightened risk of infection, especially in patients with weakened immune systems \cite{baharudin2023effect}. As another example, fluctuations in temperature within an operating room can be detrimental to the patient's well-being \cite{uscinowicz2023subjective}. Thus, providing integrity to the transmitted data is essential in the healthcare-IoT (H-IoT) environment. Any disruption can adversely affect patient care, including delayed or prematurely accelerated data caused by cyber threats \cite{khatun2023machine}, reducing potential disruptions that could adversely affect clinical results.

This research investigates the accuracy of convolutional neural network (CNN) models in detecting anomalies in time series data and their adaptability to real-world constraints within healthcare-IoT, as simulated within the Cooja environment. CNN is widely used for various applications, including the analysis of time series data, for which it can achieve competitive and better results than traditional time series analysis models like Autoregressive Integrated Moving Average (ARIMA) \cite{wang2024robust}.

The primary goal is to detect anomalous readings on environmental sensors within the hospital's IoT ecosystem. Moreover, our dataset improves abnormal reading detection, reduces risks, enhances efficiency, and elevates healthcare quality in H-IoT environments.

\section{Methodology} \label{sec:methodology}
This paper proposes a novel CNN-based method for identifying time-series healthcare-IoT data anomalies. The methodology is explicitly tailored to the healthcare-IoT environment. Models and data are analyzed using Python 3.9, integrated with a neural network framework. To ensure robust and efficient deep learning processing, all experiments are conducted on a \textit{MacBook Pro} equipped with the \textit{M1 chip}. The performance of the algorithms has been measured using our developed dataset ``WSN\textunderscore DDoS\textunderscore Attack\textunderscore H-IoT2023" in this research.

\subsection{Data Acquisition}
To detect time series anomalies, we divide the healthcare environment into specific zones such as patient wards, waiting rooms, medical equipment areas, and high-traffic areas. 


\subsection{Data Preprocessing}
The first step is establishing a baseline profile for generating datasets from healthcare devices' raw data packets. The baseline profile consists of three features, namely \textbf{Time, Mote, and Message}. 


\textbf{Feature Extraction:} Several features are extracted from the raw data descriptions through pattern matching, including `Previous Send Time,' `Interval,' `Previous Send Interval (Same/current Node),' etc. The goal is to increase accuracy by experimenting with five selected features \cite{9038563}. 



\begin{itemize}

    \item \textbf{IDs:} In the Cooja simulation, IDs serve as unique identifiers for nodes and servers, facilitating efficient tracking and management of client-server interactions. Furthermore, a unique identifier for a device, node, mote, or client.  

    \item \textbf{Interval:} A period between two or more data transmissions. The equation for finding the interval is shown in equation \eqref{eq:interval}. 
        \begin{equation}
        \label{eq:interval}
        \text{Interval} = \text{TimeStamp} - \text{PreviousSendTime}
        \end{equation}

   \item \textbf{Previous Send Interval (\textbf{Same Node}):} Approximately one minute has passed since the last transmission of data from the current node. In equation \eqref{eq:PSIl}, ``PSI" stands for ``PreviousSendInterval," ``TS" stands for ``TimeStamp," and ``PST" stands for ``PreviousSendTime", calculated as, 
            \begin{equation}
            \label{eq:PSIl}
            \begin{aligned}
            \text{PSI(same node)} = \text{TS} - \text{PST(same node)}
            \end{aligned}
            \end{equation}

 
    \item \textbf{Average Previous Send Interval (Same Node):} The Average Previous send interval (same node) is calculated based on the Previous send interval (same node). In this equation \eqref{eq:average}, ``Avg. PSI" stands for ``Average Previous Send Interval," ``Sum of PSI" represents the sum of all previous send intervals, and ``Total Intervals" signifies the total number of intervals for the same node (SM). The calculation is as follows:
            \begin{equation}
            \label{eq:average}
            \text{Avg. PSI (SM)} = \frac{\sum \text{PSI(SM)}}{\text{Total Intervals (SM)}}
            \end{equation}
            
   \item \textbf{Previous Sender:} Before the current sensor node, a node transmits data, for example, sensor node ID:8 and the current node ID:2. 

\end{itemize}

\textbf{Data Analysis:} To capture average behaviors across different nodes, features such as `Average Previous Send Interval (Same Node), `Average Previous Send Interval (Normal Nodes), and `Average Previous Send Interval (Malicious Nodes)' are computed.




\textbf{Time Series Data Splitting:} When handling time series data, paying attention to the temporal structure is crucial. A time series dataset differs from a standard dataset in that data points are arranged sequentially instead of independently. A data sequence is essential for identifying underlying patterns and relationships.
When splitting, a specialized method is used to preserve the temporal structure. This approach divides the dataset based on sequential order rather than random splits. In particular, the first part of the set (e.g., 70\%) is used for training, while the remaining portion is used for testing.
As a result, any potential data leakage is eliminated since the training dataset always precedes the test dataset sequentially. It is, therefore, possible to predict tasks that depend on past events by maintaining the natural order of the data.





\subsection{CNN Model Architecture}
A time series CNN architecture is explicitly developed for one-dimensional time series data. A 3D array is created by reshaping an input into samples, timesteps, and individual features. Following this, a convolutional layer is created using 32 filters of size three and a Rectified Linear Unit (ReLU) activation function, which excels at detecting temporal patterns. Following that, a max-pooling layer with a pool size of 2 reduces the dimensionality of the data. In binary classification, the dense layer is initially flattened. The input is passed through a Sigmoid activation function in the next dense layer.



The sigmoid activation for the dense layer is shown below:

\[ \sigma(z) = \frac{1}{1 + e^{-z}} \]

As compared to Softmax, ReLU, and Tanh, Sigmoid performed better in our binary classification experiment. Table \ref{tab:activation_scores} indicates this. In binary classification, Sigmoid is preferred due to its effective outcome mapping. Softmax, Tanh, and ReLU comparisons also highlight the importance of carefully selecting activation functions based on data and task characteristics. 

\begin{table}[h]
\centering
\begin{tabular}{ccc}
\hline
Activation Function & Accuracy (\%) & Estimated Loss (\%) \\ \hline
Sigmoid & 92\% & 8\% \\
Softmax & 47\% & 53\% \\
ReLU & 68\% & 32\% \\
Tanh & 76\% & 24\% \\ \hline
\end{tabular}
\caption{Scores with Different Activation Functions}
\label{tab:activation_scores}
\end{table}




\subsection{Experimental Setup}
The simulations are conducted using the Cooja simulator on Contiki 3.0 OS. This version contains several improvements, including enhancements to Cooja, improvements to MSPSim, new regression tests, new platforms, enhancements to the radio API, enhancements to the IPv6 stack, implementation of the message queuing telemetry transport (MQTT) protocol, updates to the constrained application protocol (CoAP), and added hypertext transfer protocol (HTTP) sockets \cite{ContikiOS2023}. The latest version of Contiki comes with all the tools and compilers necessary for emulating hardware devices and debugging programs. A free, open-source platform is available for download and testing anytime. Cooja has two emulator software packages: Avrora and MSPSim \cite{fahmy2023simulators}. As a result, it emulates MSP430-based devices through MSPSim. Most platforms are equipped with the MSP430 microcontroller \cite{rouissat2022potential}.
Consequently, MSPSim is the most widely used software package for simulating WSNs. Cooja can emulate multiple platforms, including TelosB/SkyMote, Zolertia Z1 mote, Wismote, ESB, and MicaZ mote. In this work, we rely on the SkyMote platform. This allowed us to evaluate code and systems before their deployment on real hardware. Also, efficient node power estimation and radio transmission display are ensured.


The technical specifications of the system used during the simulation stage are as follows:

\begin{itemize}
    \item \textbf{CPU:} 12th Gen Intel(R) Core(TM) i5-12500H   2.50 GHz
    \item \textbf{Random Access Memory:} 16 GB
    \item \textbf{Operating System:} Windows 11
    \item \textbf{GPU:} NVIDIA GeForce RTX
    \item \textbf{Emulated Environment:} Oracle VM VirtualBox 7.0 Platform
    \item \textbf{Simulation Spotlight:} Contiki 3.0 OS  
\end{itemize}
Integrating these hardware characteristics with Oracle VM VirtualBox 7.0 creates a flexible and efficient platform. This setup makes detecting abnormalities and significant events from IoT devices in healthcare easier.

\subsection{Metrics for Performance Evaluation}
This section illustrates the level of precision of the suggested classification method and compares its performance to that of established methods, which employ a CNN classifier. The accuracy is calculated as,

\begin{equation} 
 \text{Accuracy}   =  \frac{Correctly \ classified \ nodes}{Total\ nodes\ in\ test\ data}*100\%  \label{i}  
\end{equation}

Thus, the error rate is calculated in equation  
\begin{equation} 
 \text{Error\ Rate}  =  \big(100\ -\ Accuracy) * 100\%  \label{ii}
 \end{equation}


\subsection{Convolutional Neural Network Classifier}
This paper relies on CNN as the basis for classification. The proposed model extracts features from input data using a \textit{1D convolutional layer}, particularly suited for time-series analysis. The initial layer consists of \textit{32 filters}, each using a three-unit kernel, optimizing the model's performance in isolating time-related attributes. A \textit{`relu'} activation function is incorporated into the convolutional framework to enhance the model's ability to recognize complex relationships in data. As a result of this particular function, the model can distinguish complex features and anomalies more easily. Following the convolution stage, the data is strategically streamlined using 
\textit{MaxPooling1D}.

\begin{figure} [ht!]
    \centering
    \includegraphics[width= 0.5 \textwidth]{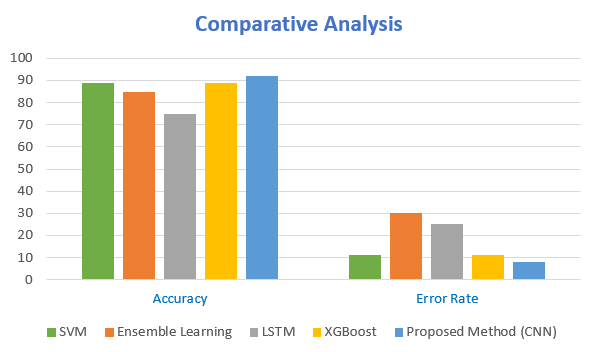}
    \caption{Comparative Performance: Accuracy vs. Error Rate}
    \label{fig:accuracy}
\end{figure}

In our experiments, all algorithms displayed in the graph \ref{fig:accuracy} are accurate to a reasonable degree. Nevertheless, the decision leaned towards the CNN model for various reasons. CNN not only demonstrated faster training times but also required fewer computational resources. It outperforms other algorithms by extracting relevant features from our dataset. In addition, CNN consistently performs better than other methods, such as support vector machine (SVM) and ensemble learning, regarding accuracy and loss metrics. The following list summarizes the execution times for CNN and SVM:

\begin{itemize}
    \item \textbf{CNN (10 epochs):} 1 minute 40 seconds
    \item \textbf{SVM:} 2 minutes 17 seconds
\end{itemize}

Based on these aspects, we conclude that the CNN model is the most effective and efficient.

\section{Conclusion} \label{sec:conclusion}
This research presents an innovative method for identifying anomalies in time series data using CNN for environmental IoT sensors in healthcare. This paper describes a unique dataset called ``WSN\textunderscore DDoS\textunderscore Attack\textunderscore H-IoT2023". As a result of methodically assembling data on room temperature and humidity, this dataset reflects the commitment to cater to the unique requirements of healthcare environments. The CNN model achieved a 92\% accuracy rate on this dataset, performing better than other methods such as SVM, ensemble learning, long short-term memory network (LSTM), etc. 


\section*{Acknowledgment}
This publication has emanated from research conducted with the financial support of Science Foundation Ireland under Grant number 18/CRT/6049.

\bibliographystyle{IEEEtran}
\bibliography{IEEEabrv,references}

\end{document}